\documentclass[10pt, a4paper]{article}
\usepackage{lrec2022} 
\usepackage{multibib}
\newcites{languageresource}{Language Resources}
\usepackage{graphicx}
\usepackage{tabularx}
\usepackage{soul}
\usepackage{mathtools}
\usepackage{amssymb}
\usepackage{mathrsfs}
\usepackage{booktabs}
\usepackage{multirow}
\usepackage{titlesec}
\titleformat{\section}{\normalfont\large\bfseries\center}{\thesection.}{1em}{}
\titleformat{\subsection}{\normalfont\SmallTitleFont\bfseries\raggedright}{\thesubsection.}{1em}{}
\titleformat{\subsubsection}{\normalfont\normalsize\bfseries\raggedright}{\thesubsubsection.}{1em}{}
\renewcommand\thesection{\arabic{section}}
\renewcommand\thesubsection{\thesection.\arabic{subsection}}
\renewcommand\thesubsubsection{\thesubsection.\arabic{subsubsection}}

\usepackage{epstopdf}
\usepackage[utf8]{inputenc}

\usepackage{hyperref}
\usepackage{xstring}
\usepackage{amsmath}

\usepackage{color}

\title{Towards a Cleaner Document-Oriented Multilingual Crawled Corpus}

\name{Julien Abadji$^1$, Pedro Ortiz Suarez$^{1,2}$, Laurent Romary$^1$, Benoît Sagot$^1$} 

\address{Inria$^1$, Sorbonne Universit\'e$^2$.\\
         2 rue Simone Iff, 75012 Paris$^1$, 21 rue de l’École de médecine, 75006 Paris$^2$.\\
         \texttt{\{julien.abadji, pedro.ortiz, laurent.romary, benoit.sagot\}@inria.fr}\\}

\abstract{
The need for raw large raw corpora has dramatically increased in recent years with the introduction of transfer learning and semi-supervised learning methods to Natural Language Processing. And while there have been some recent attempts to manually curate the amount of data necessary to train large language models, the main way to obtain this data is still through automatic web crawling. In this paper we take the existing multilingual web corpus OSCAR and its pipeline Ungoliant that extracts and classifies data from Common Crawl at the line level, and propose a set of improvements and automatic annotations in order to produce a new document-oriented version of OSCAR that could prove more suitable to pre-train large generative language models as well as hopefully other applications in Natural Language Processing and Digital Humanities.  
 \\ \newline  }

\begin{document}

\maketitleabstract

\section{Introduction}

The demand for large corpora has considerably increased in recent years with the advent of semi-supervised learning methods in Natural Language Processing (NLP), such as \emph{word embeddings} \cite{mikolov-etal-2013-distributed,pennington-etal-2014-glove,mikolov-etal-2018-advances}, \emph{contextualized word representations} \cite{howard-ruder-2018-universal,peters-etal-2018-deep,devlin-etal-2019-bert} and more recently \emph{very large generative language models} like GPT-3, T5, GPT-Neo \cite{raffel-etal-2020-exploring,brown-etal-2020-language,black-etal-2021-gpt}. While there have been some recent efforts to manually curate such corpora\footnote{\url{https://bigscience.huggingface.co}} \cite{gao-etal-2020-pile}, the common approach to collect large amounts of raw textual data still relies primarily on crawled web text \cite{ortiz-suarez-etal-2019-asynchronous,ortiz-suarez-etal-2020-monolingual,xue-etal-2021-mt5,el-kishky-etal-2020-ccaligned,espla-etal-2019-paracrawl,banon-etal-2020-paracrawl,gao-etal-2020-pile}, and although some of the initial concerns of using crawled data \cite{trinh-le-2018-a,radford-etal-2019-language} have been addressed in recent years \cite{ortiz-suarez-etal-2020-monolingual,martin-etal-2020-camembert} there a many concerns that still need to be tackled \cite{caswell-etal-2020-language} specially for multilingual data \cite{caswell-etal-2021-quality}.

In this demand for large raw textual corpora we can observe a clear back and forth in the type of data used to pre-train these models. On one hand some authors have opted for highly curated or edited data like Wikipedia such as \newcite{al-rfou-etal-2013-polyglot} and \newcite{bojanowski-etal-2017-enriching} for static word embeddings, the 1B Word Benchmark \cite{chelba-etal-2014-one} for ELMo \cite{peters-etal-2018-deep}, and the BookCorpus \cite{zhu-etal-2015-aligning} and Wikipedia for BERT \cite{devlin-etal-2019-bert}. On the other hand projects like those of \newcite{pennington-etal-2014-glove} or \newcite{grave-etal-2018-learning} used crawled data for the pre-training of fixed word embeddings, CamemBERT \cite{martin-etal-2020-camembert} a contextualized model for French successfully used only Crawled data for pre-training, and even large generative language models like T5 have used mainly crawled data successfully \cite{raffel-etal-2020-exploring}. We can of course also see examples of projects successfully using a mix of both manually curated and automatically crawled data such as RoBERTa \cite{liu-etal-2019-roberta}, XLNet \cite{yang-etal-2019-xlnet} and GPT-Neo \cite{black-etal-2021-gpt,gao-etal-2020-pile}. However, no matter the chosen approach to build these large corpora, there are in every case concerns that have been expressed, specially for the datasets used in very large generative language models \cite{bender-etal-2021-on}, even when using manually edited resources like Wikipedia \cite{barera-2020-mind}.

In this paper, that is part of the OSCAR project\footnote{\url{https://oscar-corpus.com}} or \emph{\textbf{O}pen \textbf{S}uper-large \textbf{C}rawled \textbf{A}ggregated co\textbf{R}pus} \cite{ortiz-suarez-etal-2019-asynchronous,ortiz-suarez-etal-2020-monolingual,abadji-etal-2021-ungoliant} we would like to tackle some of the existing problems with OSCAR and its pipeline \emph{Ungoliant}\footnote{\url{https://github.com/oscar-corpus/ungoliant}} pointed out by \newcite{caswell-etal-2020-language,caswell-etal-2021-quality}, by completely shifting our language classification pipeline Ungoliant from line level classification, to document level language classification. Moreover we propose a new set of automatic annotations that we add to the document metadata after language classification and that we hope will help OSCAR users more easily determine which documents they would like to use.

The contributions of the paper are as follows:

\begin{itemize}
    \item A new, document oriented corpus that is comparable in total size and language size distribution with OSCAR 21.09,
    \item A line filtering that intends to limit the integrity destruction of the documents, keeping contiguous lines and making documents human readable and exploitable as documents,
    \item Annotations that enable quality related filtering, enabling the query of documents meeting certain length criteria, potentially increasing the quality of data for less data hungry applications,
    \item A 12GB multilingual corpus,
    \item A deduplicated English corpus, as well as a line deduplication tool 
\end{itemize}

While we are aware that this set of improvements still does not address all the concerns expressed by \newcite{caswell-etal-2021-quality} or \newcite{bender-etal-2021-on}. We still believe the new proposed features as well as the release of the OSCAR 22.01 will hopefully be of use to the users of the OSCAR projects, specially considering that maintaining an up-to date, manually curated, large multilingual corpus still remains a very expensive, time-consuming task.

\section{Related Work}

Crawled data and more specifically Common Crawl\footnote{\url{https://commoncrawl.org}} has been extensively used for pre-training language representations and large generative language models in recent years. One of the first proposed pipelines to automatically classify Common Crawl by language was that of \newcite{grave-etal-2018-learning}, it classified Common Crawl entries at line level using the FastText linear classifier \cite{joulin-etal-2016-fasttext,joulin-etal-2017-bag}. However, even though FastText word embeddings were released for 157 different languages \cite{grave-etal-2018-learning}, the data itself was never released.

Later \newcite{ortiz-suarez-etal-2019-asynchronous} reproduced and optimized \newcite{grave-etal-2018-learning} pipeline and actually released the data which came to be the first version of the OSCAR corpus (now referred to as OSCAR 2019). This pipeline was then rewritten and optimized by \newcite{abadji-etal-2021-ungoliant} which in turn released a second version of OSCAR (referred to as OSCAR 21.09) but, other than adding the metadata and using a more recent dump of Common Crawl, it remained virtually the same as the original one proposed by \newcite{ortiz-suarez-etal-2019-asynchronous}. All these three mentioned pipelines \cite{grave-etal-2018-learning,ortiz-suarez-etal-2019-asynchronous,abadji-etal-2021-ungoliant} classified Common Crawl's text at the line level, meaning that the apparent \emph{``documents''} of OSCAR were actually just contiguous lines of text that were classified as being the same language. This approach preserved somehow the document integrity of monolingual entries in Common Crawl, but it completely destroyed the document integrity of multilingual entries.

Parallel to the development of OSCAR, there is also Multilingual C4 (mC4) \cite{xue-etal-2021-mt5} and CCNet \cite{wenzek-etal-2020-ccnet} both of which are also derived from Common Crawl but propose pipelines that propose a document level language classification as opposed to OSCAR's line level classification. Both CCNet and mC4 pipelines proposed methods for filtering \emph{``undesired''} data: CCNet used small language models trained on Wikipedia and based on the KenLM library \cite{heafield-2011-kenlm} while mC4 used a simple badword filter\footnote{\url{https://github.com/LDNOOBW/}}.

\section{Filtering}

Previous OSCAR pipelines were line-oriented (where a line is defined as a string separated by \texttt{\textbackslash n}), which meant that the highest filtering granularity were lines. 
Having a document-oriented corpus implies that:
\begin{itemize}
    \item We must try to keep the document integrity, by altering it in a way that does not completely destroy its coherence.
    \item Operations on the document (filtering, identification, annotation) must take into account the document as a whole.
\end{itemize}

We aim to produce a corpus that is similar in size and quality to OSCAR 21.09, looking for a set of filters that limits the inclusion of short, noisy lines in documents, while keeping a sufficient quantity of data, especially for low- and mid-resource languages. Those filters either keep/discard a given document, or remove lines from the document body then keep it.

\subsection {Header and footer filter}

Similar to previous OSCAR pipelines, we use a length-based filter discarding short-lines. However, we restrict the removal on contiguous sequences of short lines that are located either at the head or at the tail of the document. In the following document, only the lines preceded by an exclamation point would be kept.

\begin{verbatim}
Home
Login
Sign Up
Welcome to my Website
! Lorem Ipsum Dolor Sit Amet ....
! Lorem Ipsum Dolor Sit Amet ....
! Lorem Ipsum Dolor Sit Amet ....
! Lorem Ipsum Dolor Sit Amet ....
Copyright Myself
Legal
Contact
\end{verbatim}

The solution still has numerous drawbacks, especially when dealing with documents crawled from the internet, a source known to be extremely noisy and full of edge cases: Adding a long line at the very head and tail of the previous document would completely negate the benefits of the filter.

\subsection{Short lines proportion filter} 

In order to refine the filtering process, we use a count-based filter that separates the data in two bins: One for short lines and one for long lines. The filter then checks which bin is bigger, and filters out documents where the short lines bin is bigger.

This filter may limit the impact of documents containing low-quality long lines at the head/tail, then a high number of short lines.

\section{Identification}

The backbone of the language identification process is similar to the one used in goclassy \cite{ortiz-suarez-etal-2019-asynchronous} for the generation of OSCAR 2019 and Ungoliant \cite{abadji-etal-2021-ungoliant} for the generation of OSCAR 21.09. However, shifting to a document oriented corpus (with a single top-level identification per document) requires to infer the document identification, based on line identifications.

We define a document $\mathcal{D}$ as a pair $\mathcal{D}=(\mathcal{L}, \mathscr{L})$ where $\mathcal{L}=\{l_1,\ldots,l_n\}$ is the set of lines (strings separated by \texttt{\textbackslash n}) that constitute the document and $\mathscr{L} = \{g_1, \ldots, g_m\}$\footnote{Note that since FastText identifies one language by line, we have always have $m\le n$ for every document $\mathcal{D}$.} is the set of languages identified by FastText for the document $\mathcal{D}$. When FastText is no able to identify a language for an specific line, for instance because the confidence isn't higher than $0.8$, we tag said line with the \emph{No Identification Language} that we simply note by $g_0$. Furthermore, we define each line $l_i$ in a document $\mathcal{D}$ as a triplet $l_k=(g_i, p_i, s_i)$ where $g_i$ is the language identified by FastText with the highest confidence for the line $l_i$, $p_i$ is said confidence and $s_i$ is the size in bytes of the line $l_i$. We also note $|l_i|=s_i$ and we thus define the size $|\mathcal{D}|$ of a document $\mathcal{D}$ as
\[
    |\mathcal{D}| = \sum_{i=0}^{n} |l_i| = \sum_{i=0}^{n} s_i.
\]
Moreover, for each identified language $g_j \in \mathscr{L}$ in a document containing $n$ lines, we define its size $|g_j|$ as
\[
    |g_j| = \sum_{\mathclap{\{s_i \mid g_i = g_j\}}} s_i.
\]
Finally for each language $g_j \in \mathscr{L}$ we can also compute its \emph{overall weighted confidence} $P_j$ throughout the document $\mathcal{D}$ as the following weighted mean:
\[
    |P_j| = |D|^{-1}\sum_{\mathclap{\{s_i|g_i=g_j\}}} s_jp_j.
\]

\subsection{Multilingual document identification}

A document can contain lines in multiple languages for several reasons: 
\begin{enumerate}
    \item Identification mismatch, that can show up frequently, especially with languages that have significant vocabulary overlap (Czech and Slovak),
    \item Crawl from a website where the interface is written in a language, and the body is written in another one,
    \item Crawl from a translation page, where the same content is present in two (or more) different languages.
\end{enumerate}

In these examples, we should aim to limit the presence of 1. and 2., while maximizing the presence of 3.: documents having a balanced set of lines per language. Thus, we decide to take a cautious approach, restricting the multilingual document identification test to the documents that:
\begin{itemize}
    \item Have at least $5$ lines,
    \item Have at most $5$ different languages.
\end{itemize}
Next, we compute the \emph{proportion} for each language $g_j \in \mathscr{L}$ in the document $\mathcal{D}$ defined as follows
\[
    \mathrm{Pr}_g = \frac{|g|}{|\mathcal{D}|},
\]
including for the no identification language $g_0$.

A document $\mathcal{D}$ containing $n$ lines is identified as multilingual if and only if: 
\[
    \begin{dcases}
        |g_j| \ge \frac{|\mathcal{D}|}{n+1} &\forall g_j \neq g_0, \text{and}\\
        |g_0| \le \frac{|\mathcal{D}|}{n+1}        
    \end{dcases}
\]
As an example, a document holding $m=3$ languages is multilingual if each language makes up at least $\frac{1}{m+1} = \frac{1}{4}$ of the document, and that there is at most $\frac{1}{4}$ of the document that is of unknown identification.

\subsection{Monolingual identification}
We begin by identifying each line, keeping in memory the language identified, the confidence of the identification, and the size of the line. We keep track of lines that have not been identified with a special token, and a confidence of 1.

If the document does not pass the multilingual check, we then take the largest represented language and compute its overall confidence $P_j$ and use a minimum confidence threshold of $0.6$ that is way lower than the previous pipelines ($0.8$). This is motivated by the following reason: The document-based filtering removes documents containing lines that could have been kept by former pipelines, thus reducing the size of the generated data. 

Using a lower threshold could help getting lower-quality documents that still hold high-confidence lines in themselves. 

\section{Annotation}

While the filtering and identification steps are lenient by using lower thresholds than the previous pipelines, we introduce annotations, as non-destructive filters that enable more precise downstream filtering for the corpus users, as well as a useful resource to quickly assess the quality of a corpus. Annotations enable more aggressive filters to be run, since the non-destructive nature of annotations can in turn be used to refine annotation filters.

Numerous annotations are available, and each document can have several ones at the same time.

\subsection{Length-based annotations}

Some simple annotations are added when documents doesn't meet certain length requirements:

\begin{itemize}
    \item The document has a low ($\le 5$) number of lines (\emph{tiny})
    \item The document has a high number ($\ge 50\%$) of short lines (\emph{short\_sentences})
\end{itemize}

These annotations helps spotting potentially tiny documents, where the line structure or the document size could negatively influence training tasks.

A third annotation checks the occurrence of short lines at the start of the document, and adds a \emph{header} annotation if it is the case, indicating that low-quality content could be present at the start of the document.

A fourth annotation named \emph{footer} works in the same way on the tail of the document.

\subsection{Noise detection}

Some documents make their way into the corpus while being extremely noisy or non-linguistic. As an example, source code can be found in English corpora because of the presence of English words in the source itself. 

We use a filter that computes a ratio between letters and non-letters.

This filter is based on Unicode categories. We use categories \emph{Lu, Ll, Lt, Lm, Lo}\footnote{Lu: Uppercase letter, Ll: Lowercase letter, Lt: Titlecase, Lm: Modifier, Lo: Other} for letters, and we add categories \emph{Mn, Mc, Me}\footnote{Mn: Nonspacing mark, Ms: Spacing mark, Me: Enclosing mark} for accents and diacritics.

A \emph{noisy} annotation is added if the ratio passes a certain threshold, set to $0.5$.

\subsection{Adult documents}

We use the UT1 blocklist\footnote{\url{https://dsi.ut-capitole.fr/blacklists/}} as a base for adult content filtering.

The UT1 blocklist is a collection of thematic blocklists (adult, gambling, blogs, ...), usually utilized in internet access control for schools. The list is constituted and extended by both human and robots contributions (known indexes, search engines, exploration of already known addresses). The blocklist is updated twice to thrice a week by Fabrice Prigent.

Each folder contains URL and domain blocklists, enabling filtering of both websites that are centered around adult content, and websites hosting user-generated content that can be of adult nature (several social networks...).

The adult blocklist is comprised of roughly 3.7M records.







\section{Corpus}

We apply the aforementioned pipeline to the November/December 2021 crawl dump of CommonCrawl. The result is a new corpus, OSCAR 22.01. While its structure is different from the previous OSCAR corpora (due to the choice of generating a document oriented corpus), we attempt to compare the two corpora, especially in terms of size and news-related topic presence and recall. We also evaluate the occurrence and pertinence of the annotations.

\subsection{Comparison with OSCAR 21.09}
\subsubsection{Size distribution}

The data layout of OSCAR 22.01 may limit the relevance of raw size comparisons, since metadata are larger (annotations and line identifications were not present in previous OSCAR Corpora), and fused with textual data (metadata were distributed in separate files for OSCAR 21.09). 

However, comparing the distribution of corpus sizes may help us ensure that the new corpus has a size distribution similar to the older one.

\begin{figure*}[!ht]
 \begin{center}
 \includegraphics[scale=0.7]{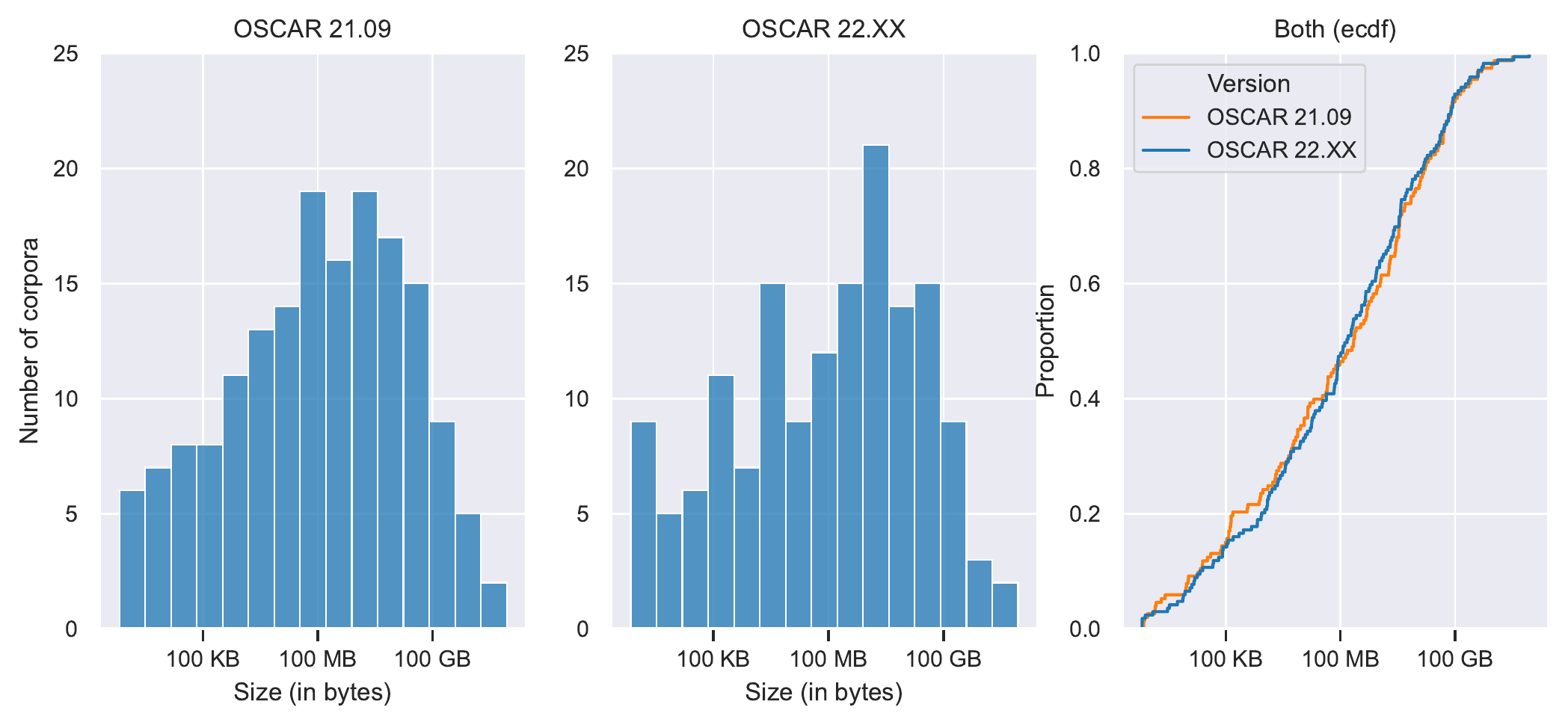} 
 \caption{Corpus size distribution between OSCAR 21.09 and 22.XX}
 \label{fig.1}
 \end{center}
 \end{figure*}

We compare the distribution of the corpus sizes between OSCAR 21.09 and OSCAR 22.01 in figure \ref{fig.1}. We see that while the overall distribution is similar, the lower end of the distribution has more variance: The $[0\text{B}, 100\text{KB})$ range shows more corpora at its bounds than at its center. We also plot the empirical cumulative density function, that helps to assert the distribution similarity between OSCAR 21.09 and OSCAR 22.01.

\begin{figure}[!ht]
 \begin{center}
 \includegraphics[width=\linewidth]{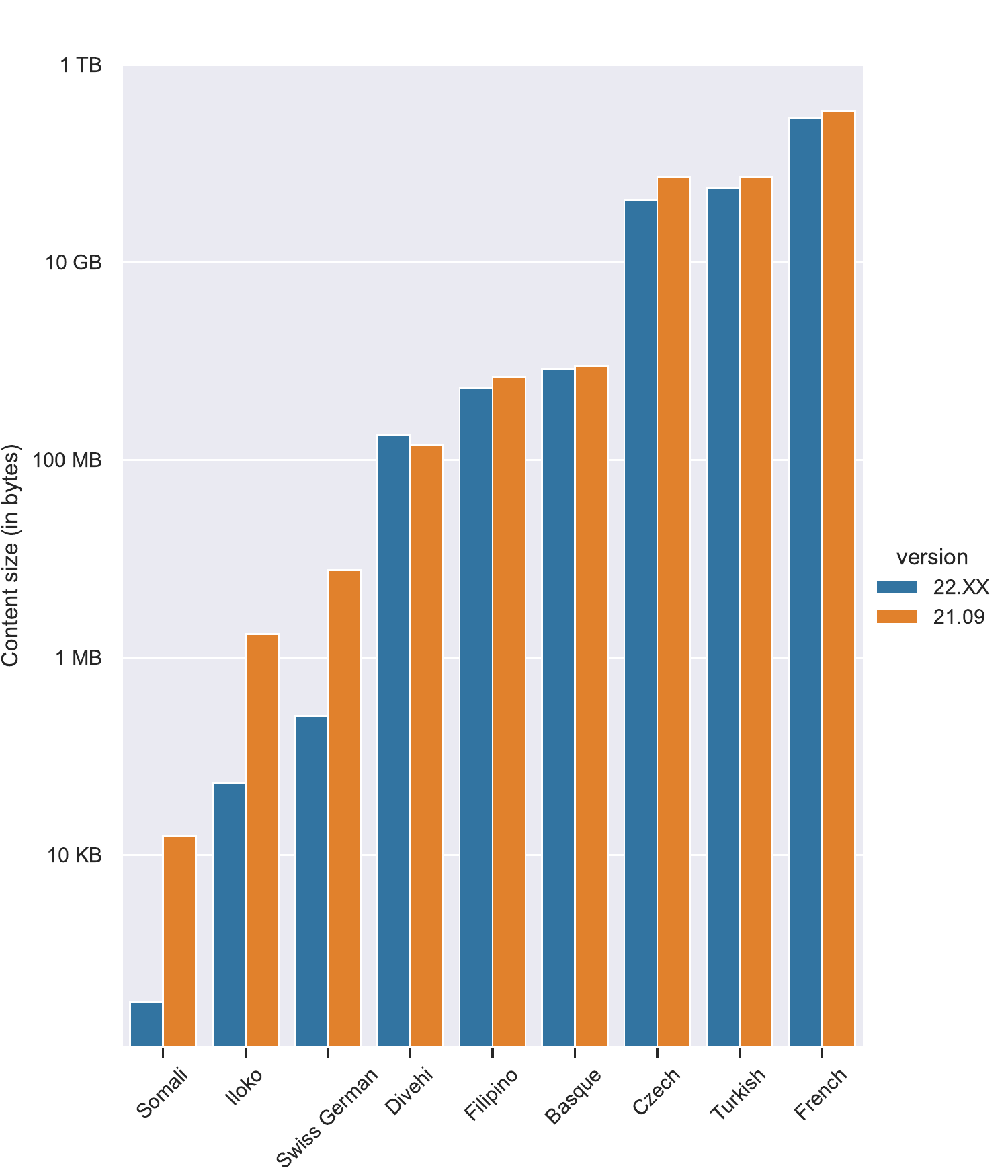} 
 \caption{Content size comparison of selected languages in OSCAR 22.01 versus OSCAR 21.09}
 \label{fig.2}
 \end{center}
 \end{figure}

 We also select three low-resourced languages, three mid-resourced languages and three high-resources languages and compare their content (that is, textual data excluding metadata) between OSCAR 22.01 and OSCAR 21.09. Comparison is shown in figure \ref{fig.2}. While the overall sizes of these corpora  have slightly decreased, the sizes of the mid and high resource languages are similar enough. 
 
\subsubsection{Size differences in low-resource languages}

The low-sized corpora exhibit important size changes. As an example, the Alemannic German corpus went from 7MB to 360KB between OSCAR 21.09 and OSCAR 22.01. This size decrease can be explained by the way the document identification works: by reasoning at a document level, documents containing a majority of German identified lines and a minority of Alemannic German identified lines will be identified as a German document, whereas previous OSCAR pipelines would have separated the lines and increase the size of the Alemannic German corpus.

By extracting the lines identified as Alemannic from the German corpus, we get around 30 MB of data, which could constitute an Alemanic corpus with a size comparable to the OSCAR 21.09 Alemanic corpus after confidence and length based filterings.

This situation can, in a way, help us investigate the cases of linguistic proximity, where languages have a lexical overlap: When a line identified as Alemannic German is found inside a document that has been identified as German:
\begin{enumerate}
    \item \label{one} Is the line in German and it is an identification error?
    \item Is the line in Alemannic German, in a document that is in German? (ex: A German website related to the Alemannic German language) 
    \item \label{three} Is the whole document in Alemannic German, and the identification classified the majority of Alemannic as German?
\end{enumerate}

Those three cases can arise and may help to enhance the detection of a said language, by finding (\ref{one}) identification mismatches, hoping that these cases would improve identification after training, or (\ref{three}), after verification by a speaker of the language, state that the whole document is in Alemannic. The new data collected could in turn be used to improve language detection.

\subsubsection{New themes}

As OSCAR 22.01 is based on a November/December 2021 dump (compared to OSCAR 21.09, based on a February 2021 dump), the corpus should include data related to events contemporary to February 2021. We conduct a simple word search similar to the one conducted for the generation of OSCAR 21.09 \cite{abadji-etal-2021-ungoliant}, using both old and new events, in order to give a rough idea of both the actuality and the memory of the corpus.

\begin{table}[t]
\centering\small
\begin{tabular}{lrrrr}
\toprule
Language & Term & 21.09 & 22.XX\\
\midrule
\multirow{1}{*}{Arabic} & Beirut port explosion & 31&13 \\
\multirow{1}{*}{Burmese*}&  Min Aung Hlaing  & 3439  & 2736\\
\multirow{1}{*}{English} & Obama & 27639    &8697  \\
\multirow{1}{*}{English} & Biden  & 19299     &8232 \\
\multirow{1}{*}{English} & Omicron  & 131     &417 \\
\multirow{1}{*}{French} & Yellow Vests& 96  &73    \\
\multirow{1}{*}{Spanish} & Aborto  & 1504     &572 \\
\bottomrule
\end{tabular}
\caption{Comparison of occurrences of news-related terms between OSCAR and our corpus in a sample of 100 CommonCrawl shards. \\ *: For the Burmese language, we use the whole 21.09 and 22.XX corpus since it is a low resource language. Terms are translated in the corpus language.}
\label{tab:word-frequency}
\end{table}

We see that the events and terms related to events predating February 2021 are still occurrent in the corpus, but have a diminished count that is in the same order of magnitude. 
We also count the occurrences of the term Omicron, related to the Omicron variant, and observe that the term has a higher count on the 21.09 sample.

\subsubsection{Absence of deduplication}

Contrary to OSCAR 21.09, we do not distribute a deduplicated version of the majority of OSCAR 22.01.

The line-level deduplication of documents would have destroyed the integrity of documents themselves, hampering human readability and even sequential sentence sense. We can imagine having forum discussions' sense destroyed because of identical responses, or song lyrics being altered.

Moreover, the similarity-based document-level deduplication procedure is very costly in terms of computing power and time \cite{gao-etal-2020-pile}. 

We make the choice of distributing a non deduplicated version of OSCAR along with a deduplicated, line oriented version of the English corpus, while encouraging the use of deduplication in the context of training language models \cite{lee-etal-2021-deduplicating}.
A line-level deduplication tool will be available as part of the OSCAR toolkit\footnote{\url{https://github.com/oscar-corpus/oscar-tools}}. We will also distribute a deduplicated version of the English part of OSCAR 22.01, with a data layout similar to OSCAR 21.09 corpora.

\subsection{Annotations}

\subsubsection{Raw stats}

Annotations helps us to infer the composition of the corpora: The \textit{tiny}, \textit{short\_sentences} and especially \textit{noisy} annotations may indicate documents of a varying poor quality, with \textit{noisy} being the worst. 

Also, comparing corpora annotation distributions, especially related to their size, could highlight potentially very low quality corpora. This semi-automated quality checking process could be used to label corpora where data quality is bad. 

We select 3 low-resource ($\simeq100KB$), 3 mid-resource ($\simeq100MB$) and 3 high-resource ($\simeq100GB$) languages and plot the number of documents per annotation, adding a \textit{total} legend for the total document count and a \textit{clean} legend for documents that do not have any annotation. We then plot the counts for each resource group using adapted scales. 

We observe that the annotation distribution is similar for each resource group, but that the lower resourced languages have a higher proportion of documents annotated with \textit{short\_sentences} and \textit{tiny}.

\begin{figure*}[!ht]
 \begin{center}
 \includegraphics[scale=0.7]{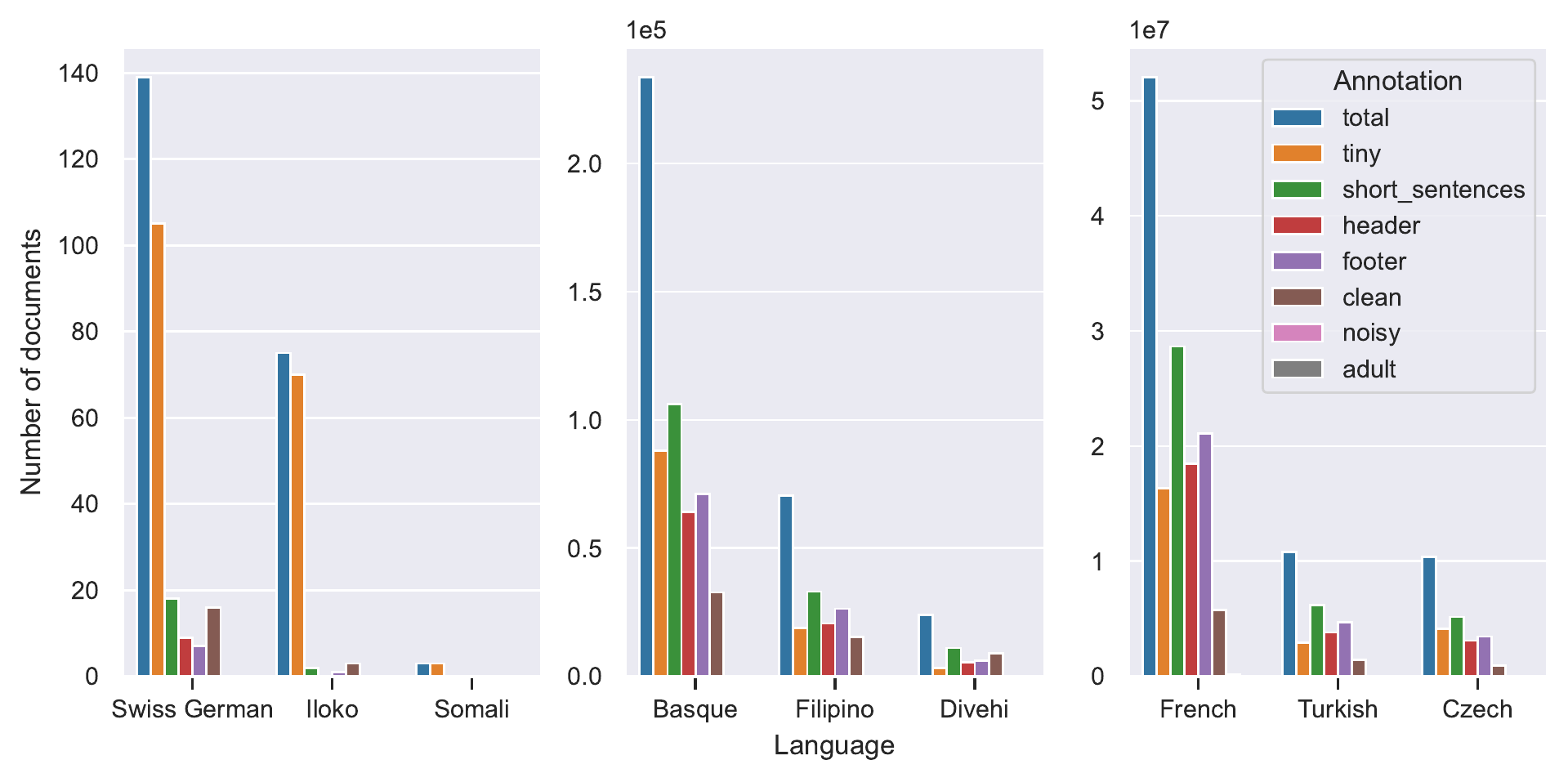} 
 \caption{Annotation count in selected low, mid and high resource languages (scales are adapted to corpus size)}
 \label{annot-count}
 \end{center}
 \end{figure*}
 
\begin{figure}[!ht]
 \begin{center}
 \includegraphics[width=\linewidth]{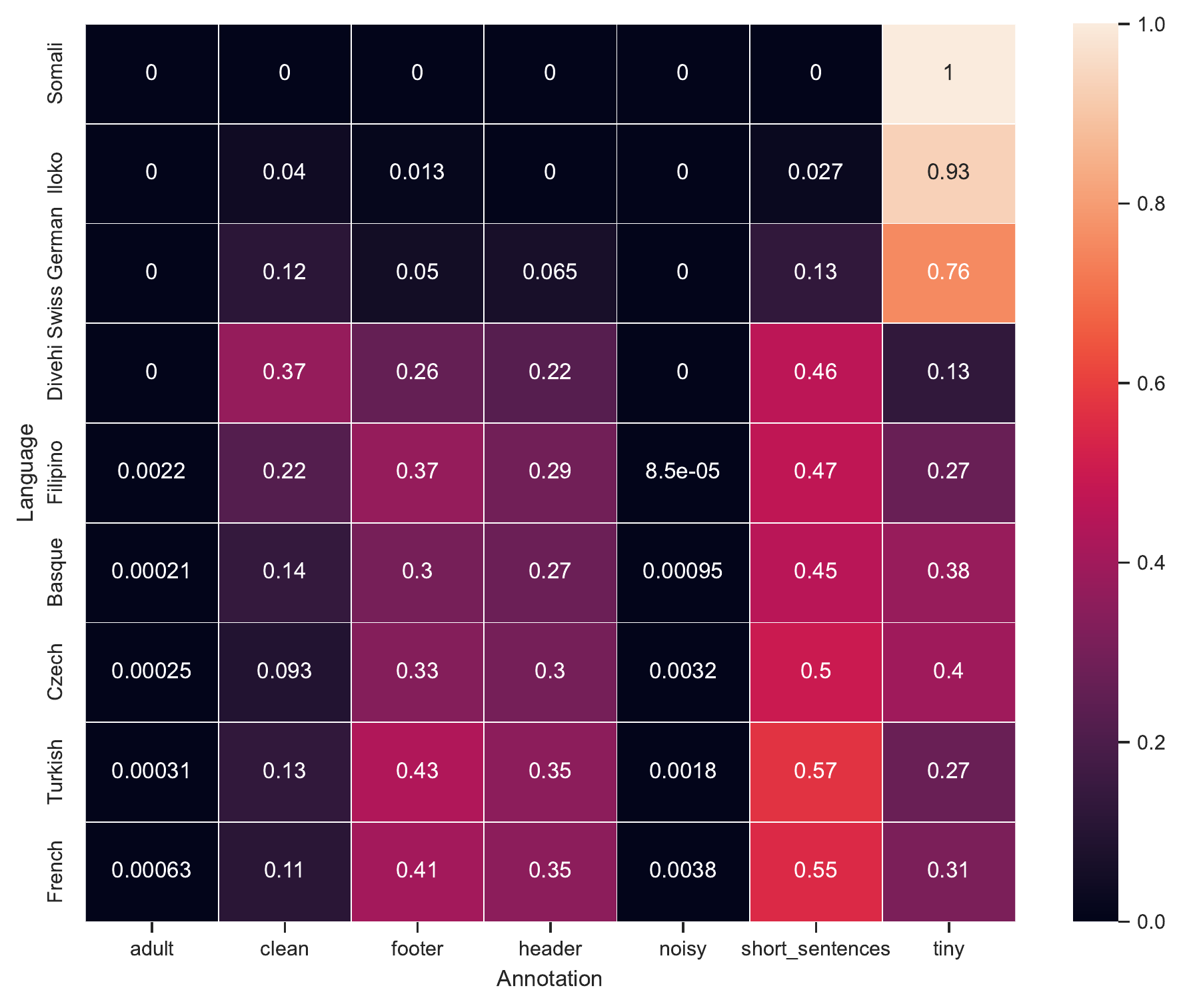} 
 \caption{Heat map of annotation distributions in selected low, mid and high resource languages.}
 \label{annot-heatmap}
 \end{center}
 \end{figure}

 In order to better compare the resource groups, we display the annotation distribution in a heat map (figure \ref{annot-heatmap}). 
 We notice important differences between low and mid/high resource groups.
 A very large proportion of the low resource group is annotated as \textit{tiny} while simultaneously detaining few documents annotated \textit{short\_sentences}, indicating the presence of long sentences within documents with a low number of sentences.
 
\subsubsection{Multilinguality}

The OSCAR 22.01 Corpus also contains a multilingual corpus, composed of documents holding lines in multiple languages. Each document contains at least 2 languages, and at most 5. 

\begin{figure}[!ht]
 \begin{center}
 \includegraphics[width=0.8\linewidth]{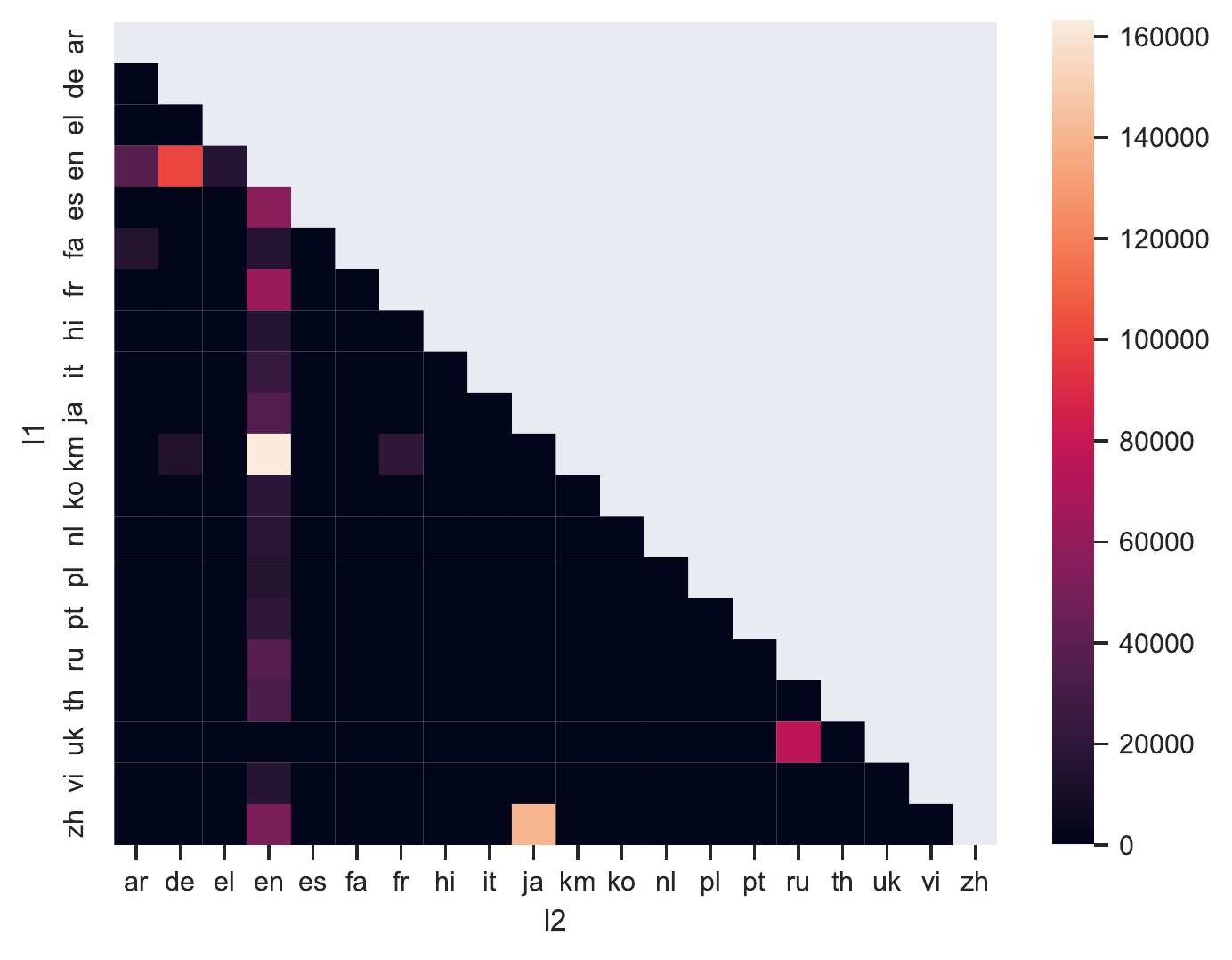} 
 \caption{Count of $(l1, l2)$ language tuples in the multilingual corpus. Languages tuples with less than 20,000 occurrences are not shown.}
 \label{multi-confusion}
 \end{center}
 \end{figure}

We check the co-occurrence of languages, highlighting the coupling of language tuples. These tuples may highlight either linguistic similarity (Czech and Slovak, Russian and Uzbek) and subsequent poor classification, errors or languages commonly found together on documents. Due to the number of languages and the sparsity of the data, we show the language couples with a number of documents greater than 20 000 (Figure \ref{multi-confusion}). 

We also note the presence of English in a high number of documents. This could be explained by boilerplate content in web pages, such as menu headers or footers.

 Using the clean annotation filter on the multilingual corpus may help to retrieve the highest quality multilingual documents.
 
\subsubsection{Clean documents}

We also look into documents that did not get annotated at all, and we find that these documents are usually of a high quality. However, their relative proportion in corpora may limit their usage.

\begin{figure}[!ht]
 \begin{center}
 \includegraphics[width=\linewidth]{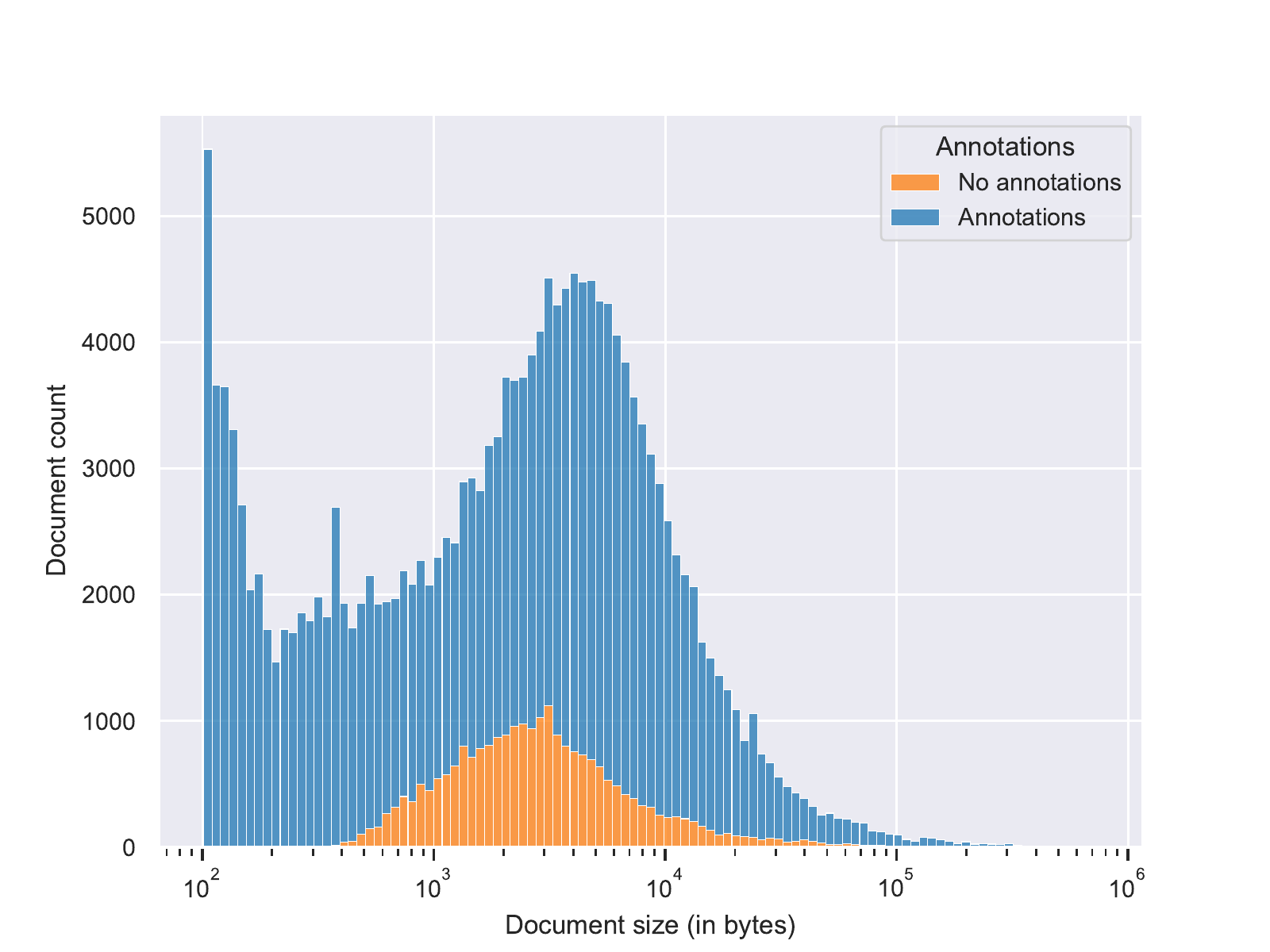} 
 \caption{Stacked distribution of annotated and non-annotated (clean) documents on a selection of the English corpus}
 \label{clean_count}
 \end{center}
 \end{figure}

We use a sample of the English corpus (183,497 documents, 1.3 GB) and compare the size of documents depending on the presence (or not) of annotations. The stacked counts are shown in figure \ref{clean_count}.

We observe that clean document mean length is slightly shorter than non-clean ones. Also, we note that while the length standard deviation of clean documents seems to be shorter, the computation yields larger numbers, caused by outliers in the high end (Annotations: $\mu=8606$ $\sigma=49874$, Clean: $\mu=6537$ $\sigma=14983$). 
By removing the top and bottom 5\%, we get (Annotations: $\mu=3686$ $\sigma=4047$, Clean: $\mu=3582$ $\sigma=3202$).

These results are not sufficient to state on the intrinsic quality of the clean documents, but may ease the study of the filters and identify future filtering needs.

\subsubsection{Adult documents}

While very small in proportions, adult annotation documents highlight interesting facts. 

The French sample contains 32,870 adult documents, out of 52,037,098. 

We count if some documents coming from tetu.com are labeled as adult, in order to probe the possibility of finding LGBTQI+ content annotated as adult. We find 1063 documents, representing $\sim 3.2\%$ of the adult documents. This may imply that more LGBTQI+ content sites are present in the blocklist, thus increasing the ratio of LGBTQI+ content labeled as adult.

We take the first 100 adult documents of the French corpus and check whether they are properly classified.
\begin{itemize}
    \item \emph{true positives} documents that exhibit explicit sexual content geared towards pornography (pornographic websites, sexually explicit fictions)
    \item \emph{false positives} documents that do not meet this criteria,
\end{itemize}

We separately count websites that are simultaneously non explicit and from LGBTQI+ websites.

We find:
\begin{itemize}
    \item $77$ true positives,
    \item $2$ false positives belonging to LGBTQI+ websites,
    \item $21$ false positives
\end{itemize}

While the majority of true positives are properly classified, numerous educational documents do appear: These type of documents exhibit an explicit language, but does feature a good document quality, and a better representation of sexuality that is less offensive compared to the usual associations between sexually explicit content and hate speech.  \cite{luccioni-viviano-2021-whats}.

The false positives are, for the majority, websites that do not belong in the blocklist in the first place. We suppose that the addresses were previously used as adult websites.





\subsubsection{Hard bounds problems}

Several pipeline steps (especially annotators), work using hard thresholds. As an example, any document that is less than 5 lines is considered to be \textit{tiny}. However, when exploring data, we can see that there is a number of documents whose number of lines is in the neighboring of the threshold, and quality is similar to the documents labeled as \textit{tiny}.

When plotting the distribution of clean and annotated corpus data, we can notice that a very high number of documents are of a tiny ($10^2 B$) size, which coincidentally happens to be the minimum size for a document to be accepted, since the first filter removes lines that are shorter than 100 characters $(\geq 10^2 B)$.

\section{Discussion}

\subsection{Corpus}
We provide a new, document-oriented corpus of the same size of OSCAR 21.09 that keeps document integrity and is easier to filter thanks to annotations.

While the mid and high resourced languages are of a similar size, several low resource languages have seen an important decrease of size. 
We still have to check whether this size decrease comes with a quality increase, since previous low resource OSCAR corpora sometimes exhibited extremely poor quality: Many non-linguistic corpora that were published and deemed unusable weeks or months after release. 

We also note that documents of similar languages could have been merged into larger corpora, and we show that the German corpus holds $\sim 30$MB of Alemannic that, with appropriate filtering, could be treated as an independent corpus. These cases of merging are also interesting to investigate, as they can explain identification mismatches and could, in turn, help to build better language identification models.
More work has to be done in order to properly map the connection between low-resource languages and mid and high resource languages potentially containing data in these languages. 

\subsection{Annotations}

The selected annotations exhibit numerous caveats that have to be addressed in the future iterations of OSCAR generation pipelines.

The length-based annotations are widespread in the corpus, especially in mid to high resource languages ($\sim50\%$ in Czech) highlighting the potential low quality of a high number of documents as well as the need of better characterizing the nature of these line length discrepancies. Web crawls often contain boilerplate content extracted from headers, footers and sidebars, and these lines are present in the Common Crawl dumps.
Another solution would be to base the whole OSCAR generation pipeline on raw HTML files, potentially multiplying the computational cost and complexity of generating corpora.

The \textit{adult} annotation, based from an adult URL blocklist, is present on a very limited set of documents. However, studies have shown that adult content has been present in a previous version of OSCAR in a larger proportion than the one measured here \cite{caswell-etal-2021-quality}, hinting at a bad performance of the blocklist based adult content filtering approach. Moreover, we noticed that the blocklist contained websites representing LGBTQI+ related topics, which damages the representation of the LGBTQI+ (association with adult content, filtering out LGBTQI+ documents, which in turn could limit the representation in downstream tasks..).
Model-based approaches may help in improving the \textit{adult} annotation, and should be the next step towards a better annotation of adult content \cite{luccioni-viviano-2021-whats}. 
\section{Bibliographical References}\label{reference}

\bibliographystyle{lrec2022-bib}
\bibliography{custom,anthology}

\appendix

\section{Carbon Footprint}\label{carbon-footprint}

Taking into consideration recent concerns regarding the power consumption and carbon footprint of machine learning experiments \cite{schwartz-etal-2020-green,bender-etal-2021-on} we report the power consumption and carbon footprint of the OSCAR generation, assuming the whole dump of Common Crawl has already been downloaded. We follow the approach of \newcite{strubell-etal-2019-energy}.

We use a single machine having 192 GB of RAM and two Intel Xeon Gold 5218 processors, which is rated at 125 W,\footnote{\href{https://ark.intel.com/content/www/us/en/ark/products/192444/intel-xeon-gold-5218-processor-22m-cache-2-30-ghz.html}{Intel Xeon Gold 5218 specification}}. For the DRAM we can use the work of \newcite{desrochers-etal-2016-a} to estimate the total power draw of 192GB of RAM at around 20W. The total power draw of this setting adds up to around 270 W.

Having this information, we can now use the formula proposed by \newcite{strubell-etal-2019-energy} in order to compute the total power required to pre-train one model from scratch:
\[
    p_t = \frac{1.58t(cp_{c} + p_r)}{1000}
\]
Where $c$ is the number of CPUs, $p_c$ is the average power draw (in Watts) from all CPU sockets and $p_r$ the average power draw from all DRAM sockets. We estimate the total power consumption by adding CPU and DRAM consumption, and then multiplying by the \emph{Power Usage Effectiveness} (PUE), which accounts for the additional energy required to support the compute infrastructure. We use a PUE coefficient of 1.58, the 2018 global average for data centers \cite{strubell-etal-2019-energy}. The total time to generate OSCAR 22.01 in this infrastructure was of 42.6 hours. We use this information to compute the total power consumption of the OSCAR generation, which amounts to 0.4266kWh.

 We can further estimate the CO\textsubscript{2} emissions in kilograms of the OSCAR generation by multiplying the total power consumption by the average CO\textsubscript{2} emissions per kWh in our region which were 38.64g/kWh in average between the 3rd and the 5th of January 2022\footnote{\href{https://www.rte-france.com/eco2mix/les-emissions-de-co2-par-kwh-produit-en-france}{Rte - éCO\textsubscript{2}mix}.}, the exact time at which the generation was run. Thus the total CO\textsubscript{2} emissions in kg for one single model can be computed as:
\[
    \text{CO}_{2}\text{e} = 0.03864 p_t
\]
Thus total CO\textsubscript{2} emissions amount to 0.01648kg or 16.48g.

\section{Language Table}

\begin{table*}[t!]
    \centering\tiny
\begin{tabular}{llll}
\toprule
                       Language &      Size &        Documents &           Words \\
\midrule
                  Afrikaans &   47.0 MB &      12,393 &       6,227,310 \\
              Tosk Albanian &  363.6 kB &         139 &          37,381 \\
                    Amharic &  461.0 MB &      37,513 &      30,481,153 \\
                  Aragonese &   10.6 kB &          12 &              51 \\
                     Arabic &   84.2 GB &   8,718,929 &   6,103,711,887 \\
            Egyptian Arabic &    2.8 MB &       1,256 &         176,096 \\
                   Assamese &  221.2 MB &      17,084 &      11,109,557 \\
                   Asturian &   73.6 kB &          77 &           3,919 \\
                     Avaric &   18.6 kB &          14 &             582 \\
                Azerbaijani &    3.5 GB &     491,847 &     291,927,692 \\
          South Azerbaijani &   14.1 MB &       5,381 &         693,746 \\
                    Bashkir &   95.5 MB &      11,198 &       5,418,474 \\
                 Belarusian &    1.8 GB &     180,046 &     107,227,860 \\
                  Bulgarian &   35.1 GB &   2,887,115 &   2,405,981,285 \\
           Bihari languages &   24.2 kB &          27 &             569 \\
                     Bangla &   15.1 GB &   1,171,501 &     751,877,226 \\
                    Tibetan &  234.5 MB &      18,683 &       2,286,269 \\
                Bishnupriya &    2.0 MB &         271 &          98,419 \\
                     Breton &   33.7 MB &      16,119 &       3,111,619 \\
                    Bosnian &   10.3 kB &          10 &             422 \\
              Russia Buriat &   32.9 kB &          39 &             785 \\
                    Catalan &   13.9 GB &   2,627,307 &   1,508,919,864 \\
                    Chechen &   14.0 MB &       4,086 &         798,766 \\
                    Cebuano &   44.6 MB &       5,742 &       5,253,785 \\
            Central Kurdish &  716.4 MB &      84,950 &      43,913,025 \\
                      Czech &   58.6 GB &  10,381,916 &   5,452,724,456 \\
                    Chuvash &   41.8 MB &       4,750 &       2,465,782 \\
                      Welsh &  409.3 MB &      90,378 &      49,488,495 \\
                     Danish &   12.6 GB &   2,265,479 &   1,454,439,292 \\
                     German &  496.7 GB &  70,075,424 &  46,826,676,844 \\
Dimli (individual language) & 706 Bytes &           1 &              19 \\
              Lower Sorbian & 707 Bytes &           1 &              17 \\
                     Divehi &  217.2 MB &      24,067 &      10,112,205 \\
                      Greek &   78.3 GB &   6,738,546 &   5,031,242,803 \\
        Emiliano-Romagnolo. & 901 Bytes &           1 &              53 \\
                    English &    3.2 TB & 431,992,659 & 377,376,402,775 \\
                  Esperanto &  558.3 MB &     111,932 &      58,416,628 \\
                    Spanish &  381.9 GB &  51,386,247 &  42,829,835,316 \\
                   Estonian &    9.2 GB &   1,362,524 &     820,975,443 \\
                     Basque &    1.1 GB &     233,658 &      97,092,942 \\
                    Persian &   77.4 GB &   7,665,871 &   6,430,164,396 \\
                    Finnish &   37.8 GB &   4,948,961 &   2,900,615,928 \\
                     French &  382.2 GB &  52,037,098 &  41,713,990,658 \\
            Western Frisian &   75.3 MB &      21,946 &       6,357,929 \\
                      Irish &   45.6 MB &      12,233 &       4,877,850 \\
            Scottish Gaelic &  137.7 kB &         136 &           7,769 \\
                   Galician &  255.2 MB &      88,803 &      27,051,212 \\
                    Guarani &    9.0 kB &          10 &             374 \\
               Goan Konkani &  787.2 kB &          46 &          38,831 \\
                   Gujarati &    4.8 GB &     136,467 &     301,170,777 \\
                     Hebrew &   30.3 GB &   3,132,396 &   2,249,377,984 \\
                      Hindi &   23.3 GB &   1,529,907 &   1,534,799,198 \\
                   Croatian &   11.2 MB &      11,462 &         505,369 \\
              Upper Sorbian &  132.8 kB &         110 &           8,825 \\
                  Hungarian &   53.9 GB &   6,866,062 &   4,598,787,907 \\
                   Armenian &    4.7 GB &     379,267 &     268,031,270 \\
                Interlingua &   40.2 kB &           6 &          10,125 \\
                 Indonesian &   17.4 GB &   2,244,622 &   1,984,195,207 \\
                      Iloko &   97.9 kB &          75 &           8,592 \\
                        Ido &   77.3 kB &         105 &           2,690 \\
                  Icelandic &    2.0 GB &     396,183 &     210,365,124 \\
                    Italian &  229.3 GB &  28,502,092 &  24,294,684,830 \\
                   Japanese &  258.7 GB &  36,328,931 &   5,592,948,356 \\
                     Lojban &    1.9 MB &         570 &         260,542 \\
                   Javanese &  152.7 kB &          70 &          10,441 \\
                   Georgian &    7.1 GB &     488,588 &     281,430,479 \\
                     Kazakh &    2.9 GB &     261,085 &     157,267,307 \\
                      Khmer &    1.9 GB &     121,910 &      30,564,131 \\
                    Kannada &    2.6 GB &     150,850 &     108,450,571 \\
                     Korean &   51.8 GB &   5,881,481 &   3,854,968,649 \\
            Karachay-Balkar &  119.6 kB &          91 &           4,089 \\
                    Kurdish &  150.3 MB &      29,906 &      17,390,759 \\
                       Komi &  119.9 kB &         127 &           3,335 \\
                    Cornish &    1.4 kB &           2 &              55 \\
                     Kyrgyz &  518.6 MB &      62,244 &      28,028,986 \\
                      Latin &    4.1 MB &       4,397 &         187,446 \\
              \bottomrule
\end{tabular}
\begin{tabular}{llll}
\toprule
                       Language &      Size &        Documents &           Words \\
\midrule
              Luxembourgish &   15.8 MB &       5,108 &       1,545,946 \\
                   Lezghian &  375.5 kB &         124 &          19,250 \\
                 Limburgish &    1.4 kB &           2 &              41 \\
                    Lombard &    2.6 kB &           2 &             225 \\
                        Lao &  337.1 MB &      28,914 &       6,682,982 \\
                 Lithuanian &   20.0 GB &   2,303,070 &   1,712,802,056 \\
                    Latvian &    8.2 GB &   1,032,987 &     707,361,898 \\
                   Maithili &   21.6 kB &          23 &             483 \\
                   Malagasy &   57.3 MB &       3,028 &       7,279,056 \\
               Eastern Mari &   11.3 MB &       1,612 &         641,525 \\
                Minangkabau &    6.0 MB &         585 &         614,613 \\
                 Macedonian &    3.6 GB &     341,775 &     244,058,579 \\
                  Malayalam &    4.1 GB &     250,972 &     137,831,247 \\
                  Mongolian &    2.8 GB &     237,719 &     176,405,432 \\
                    Marathi &    3.3 GB &     250,376 &     160,179,233 \\
               Western Mari &  743.5 kB &         155 &          43,916 \\
                      Malay &    5.3 MB &       5,228 &         217,818 \\
                    Maltese &    2.5 MB &       2,208 &         118,190 \\
               Multilingual &   12.1 GB &   1,210,685 &     936,187,711 \\
                    Burmese &    1.9 GB &     158,733 &      44,835,970 \\
                Mazanderani &  128.2 kB &          76 &           7,337 \\
          Nahuatl languages &    8.7 kB &          12 &             179 \\
                 Low German &    9.0 MB &       1,938 &       1,012,561 \\
                     Nepali &    3.7 GB &     391,947 &     177,885,116 \\
                     Newari &    5.7 MB &       1,134 &         273,837 \\
                      Dutch &  114.0 GB &  20,206,532 &  12,329,127,151 \\
          Norwegian Nynorsk &    6.8 MB &       5,835 &         459,183 \\
                  Norwegian &    2.8 GB &     973,188 &     279,182,902 \\
                    Occitan &    2.1 MB &         373 &          31,061 \\
                       Odia &  487.9 MB &      52,942 &      23,755,902 \\
                    Ossetic &   13.9 MB &       3,560 &         800,430 \\
                    Punjabi &    1.1 GB &      68,094 &      70,068,604 \\
                     Polish &  139.0 GB &  19,301,137 &  12,584,498,906 \\
                Piedmontese &    1.7 MB &         698 &         188,270 \\
            Western Panjabi &   46.7 MB &       6,790 &       4,060,419 \\
                     Pashto &  490.3 MB &      50,312 &      46,293,249 \\
                 Portuguese &  170.3 GB &  23,735,707 &  18,441,864,893 \\
                    Quechua & 744 Bytes &           1 &              14 \\
                   Romanian &   49.2 GB &   4,624,764 &   5,261,803,995 \\
                    Russian &    1.1 TB &  76,060,844 &  62,811,122,663 \\
                   Sanskrit &  136.0 MB &       4,472 &       5,671,369 \\
                      Sakha &   65.6 MB &       6,284 &       3,473,813 \\
                   Sicilian &    1.5 kB &           2 &              50 \\
                     Sindhi &  117.1 MB &      15,516 &      10,685,611 \\
            Serbian (Latin) &  931.8 kB &         738 &          92,875 \\
                    Sinhala &    2.0 GB &     108,593 &     113,179,741 \\
                     Slovak &   16.5 GB &   2,409,555 &   1,619,121,944 \\
                  Slovenian &    1.2 GB &     351,894 &     118,400,246 \\
                     Somali &    2.1 kB &           3 &             109 \\
                   Albanian &    3.0 GB &     437,287 &     326,325,149 \\
                    Serbian &    6.9 GB &     577,472 &     482,932,670 \\
                  Sundanese &    5.0 MB &         263 &         547,145 \\
                    Swedish &   48.0 GB &   7,541,278 &   5,078,331,128 \\
                    Swahili &    1.3 MB &         462 &         123,050 \\
                      Tamil &   11.4 GB &     556,772 &     452,343,748 \\
                     Telugu &    3.4 GB &     249,756 &     137,752,065 \\
                      Tajik &  870.9 MB &      46,366 &      56,627,727 \\
                       Thai &   66.1 GB &   5,030,254 &   1,626,779,846 \\
                    Turkmen &    4.4 MB &       2,485 &         276,632 \\
                   Filipino &  646.5 MB &      70,394 &      81,881,278 \\
                    Turkish &   75.1 GB &  10,826,031 &   6,421,221,358 \\
                      Tatar &  915.3 MB &      76,398 &      51,875,265 \\
                     Uyghur &  201.9 MB &      18,556 &      11,240,889 \\
                  Ukrainian &   48.8 GB &   4,558,214 &   2,879,585,992 \\
                       Urdu &    3.4 GB &     336,994 &     332,816,354 \\
                      Uzbek &   19.9 MB &       9,526 &       1,370,842 \\
                 Vietnamese &   98.9 GB &   9,587,233 &  12,283,185,482 \\
                    Volapük &  825.9 kB &         661 &          57,039 \\
                    Walloon &  105.7 kB &         138 &           4,386 \\
                      Waray &    7.6 MB &         933 &         830,872 \\
                 Wu Chinese &  137.2 kB &          88 &           3,056 \\
                     Kalmyk &    9.3 kB &           9 &             250 \\
                 Mingrelian &    7.6 MB &       2,550 &         253,333 \\
                    Yiddish &  232.5 MB &      23,418 &      15,809,780 \\
                     Yoruba &   24.7 kB &          26 &           1,042 \\
                    Chinese &  900.9 GB &  56,524,518 &  23,149,203,886 \\
\bottomrule
\end{tabular}
 \caption{Size of the OSCAR corpus by language measured in bytes and number of words. Standard UNIX human-readable notation is used for the size in byte. We define ``words'' as spaced separated tokens, which gives a good estimate of the size of each corpus for languages using Latin or Cyrillic alphabets, but might give a misleading size for other languages such as Chinese or Japanese.} 
    \label{tab:Langs}
\end{table*}

\end{document}